\documentclass{article} \pdfpagewidth=8.5in
\usepackage{ijcai18}

\usepackage{pdfpages}

 \usepackage{xcolor} \usepackage{soul} \usepackage[utf8]{inputenc} \usepackage[small]{caption}

\usepackage{amsmath} \usepackage{amstext} \usepackage{amsbsy} \usepackage{prettyref} \usepackage{latexsym} \usepackage{amssymb} \usepackage{graphicx} \usepackage{hyperref} \usepackage{braket} %\usepackage{aas_macros}

\title{Omega: An Architecture for AI Unification}

\author{Eray \"Ozkural\\  
  Celestial Intellect Cybernetics\\
  celestialintellect.com
}

\begin{document}

\maketitle

\begin{abstract}
  We introduce the open-ended, modular, self-improving Omega AI
  unification architecture which is a refinement of Solomonoff's Alpha
  architecture, as considered from first principles. The architecture
  embodies several crucial principles of general intelligence
  including diversity of representations, diversity of data types,
  integrated memory, modularity, and higher-order cognition.  We
  retain the basic design of a fundamental algorithmic substrate
  called an ``AI kernel'' for problem solving and basic cognitive
  functions like memory, and a larger, modular architecture that
  re-uses the kernel in many ways.  Omega includes eight
  representation languages and six classes of neural networks, 
  which are briefly introduced. 
  The architecture is intended to initially address data
  science automation, hence it includes many problem solving methods
  for statistical tasks.
  We review
  the broad software architecture, higher-order cognition,
  self-improvement, modular neural architectures, intelligent agents,
  the process and memory hierarchy, hardware abstraction, peer-to-peer
  computing, and data abstraction facility.
  % We review a use-case of cloud robotics to explain how the
  % architecture will function in a non-trivial setting.
\end{abstract}

\section{Introduction and Motivation}

In today's AI research, most researchers focus on specific
application problems and they develop the capabilities of their AI
solutions only to the extent that these specific applications require
them. While challenging AI problems such as natural language
understanding require a broader view, most researchers do not begin
with an all-encompassing architecture and then adapt to a specific
application. It is usually more efficient to pursue a bottom-up
development methodology for the experimental results, and as a result,
progress in ambitious architectures for generality may have
stalled.

To achieve generality, a rigorous architectural approach has several
benefits such as easing development, allowing future extensions while
remaining backwards compatible, and exposing problems before they
happen since we can conceptualize complex use-cases. In other words,
it is at least better software engineering, however, there are also
scientific benefits such as understanding the functions and
capabilities required by a general-purpose AI system much better, and
address these problems fully. Since the most general problem is
attacked, the architecture can follow a rigorous design process which
will eliminate redundancies, leading us to a more mathematically
elegant design. And finally, since use-cases will lead the design, the
result will be empirically firmer than a special-purpose application.

A design from first principles is rarely undertaken, and it is
arduous, but it can produce highly effective systems. We build upon
the most powerful architectures for general AI, and then identify the
requirements, from which we introduce refinements to the existing
architectures, introducing new architectural ideas and incorporating
new AI technologies in the process. The resulting deep technological
integration architecture is a compact, scalable, portable, AI platform
for general-purpose AI with many possible applications in wide
domains.

\section{Design Principles for Generality}

In this section, we review the requirements of a general AI system,
and from this vantage point we formulate design principles for
constructing a general system.

A general AI system cannot contain any and all specific solutions in
its memory, therefore it must equal the computer scientist in terms of
its productive capacity of solutions. The requirement of a universal
problem solver therefore is fundamental to any such design. Naturally,
this implies the existence of Turing-complete programming languages,
and a universal method to generalize -- which implies a universal
principle of induction such as Solomonoff induction. A suitably
general probabilistic inference method such as Bayesian inference is
implied since most AI problems are probabilistic in nature. It must
have practically effective training methods for learning tasks, such
as the GPU accelerated training methods used in deep learning. The
system must have an integrated memory for cumulative learning. The
architecture must be modular for better scalability and extensibility;
human brain is a little like that as the neocortex has a grid of
cortical columns, which are apparently functionally equivalent
structures.

A general AI system must be able to support robotics, however, it
should not be limited to agent architectures; it must also support
traditional applications like databases, web search, and mobile
computing. To accommodate for such a wide variety of functions, the
architecture must expose a Swiss army knife like AI toolkit, to
provide a unified AI API to developers.  Such an API can then be
served over the cloud, or via fog computing.  Machine learning
applications generally require hardware with high performance
computing support. Therefore, the architecture should be compatible
with high performance computing hardware such as GPUs, and FPGAs to be
able to scale to many clients.

The general AI system must also address all the hard challenges of a
natural environment as formulated by \cite[Chapter
2]{russell2016artificial}: the system must cope with the partially
observable environments, multi-agent environments, competition and
co-operation, stochastic environments, uncertainty, nondeterminism,
sequential environments, dynamic environments, continuous
environments, and unknown environments. A tall order, if there were
ever one. Therefore, the system must be designed with these features
of the environment in mind, for accommodating their needs.

AIXI \cite{hutter-aixigentle} addresses partially observable
environments, however, the rest of the features require architectural
support in most cases, such as the necessity of providing a
theory-theory module (a cognitive module that has a theory of other
minds), or showing that the system will discover and adapt to other
minds. To provide for multi-agent environments, the system can offer a
self-simulation virtualization layer so that the agent can conceive of
situations involving entities like itself.  To support proper modeling
of environments like with stochastic and uncertainty, we need an
extensive probabilistic representation language to deal with
non-trivial probabilistic problems; the language must cover common
models such as hiearchical hidden markov layer models; it should offer
a wide range of primitives to choose from, which must be supplied by
the architecture. The representation language must also provide the
means to combine primitives meaningfully, and obtain short programs
for common patterns. The mystique art of designing compact
representation languages therefore remains a vital part of AI
research. To provide for effective representation of things like
sequential, dynamic, continuous environments, the architecture can
provide effective representation primitives and schemas. For dealing
with unknown environments, the architecture can provide an agent
architecture that can engage in the exploration of the unknown, much
as an animal does.

Without doubt, the system must also accommodate common data types, and
common tasks such as speech recognition, and the examples for more
specific operations should be provided. It is important that the
system allows one to implement a wide family of AI tasks for the
system to be considered sufficiently general. If, for instance, the
user cannot feasibly implement something like style transfer, that is
popular in deep learning research, with the architecture, it should
rather not be termed general. The system should support a wide range
of structured, and unstructured data, including popular data types
like image, audio, video, speech, and text, and have sufficiently rich
models to represent these challenging kinds of data.  These more human
data types constitute the primary means by which humans can
communicate with AI's directly. However, structured, regular and
irregular data types also must be supported, since these originate
from a variety of sources that can be consumed by the AI system.

The system must also therefore provide an adequate perception
architecture by which such a system can learn a world-representation
from its sensorium which includes many senses. These processes should
be sufficiently general that they can be adapted to any sort of
sensorium that will work under known laws of physics. The system
should also support an adequate intelligent agent architecture that
supports typical goal following, or utility maximization
architectures.

Therefore, it also is a challenge to test system
generality. Typically, a benchmark that consists of a large number of
diverse AI tasks and datasets must be provided for the system to
demonstrate generality. The benchmark should be diverse enough to
include the whole gamut of AI problems such as typical pattern
recognition problems of image recognition, speech recognition, but
also natural language understanding, machine learning tasks like
anomaly detection (over real-world datasets such as 
an industrial dataset),
time-series prediction (commonly used for stock market analysis),
robotics problems, game playing problems, and so forth with randomly
varied parameters.

\begin{figure}
  \centering
  \includegraphics[trim=0cm 6cm 0cm 4cm, clip=true,
  scale=0.35]{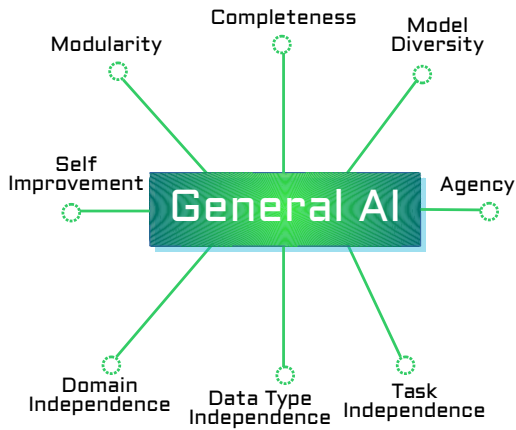}
  \caption{Some principles of general intelligence.}
  \label{fig:general-principles}
\end{figure}

We therefore arrive at an understanding of general-purpose AI design
that tries to maximize generality for every distinct aspect of a
problem. The solution space must be wide enough to cover every
problem domain. The methods must be independent from the data
type. The tasks that can be performed should not be fixed. The system
should be independent from the task to be solved; any task should be
specifiable. The architecture must not depend either on a particular
representation, it should cover a very wide range of representations
to be able to deal with different kinds of environments. The
intelligent agent code should not be environment specific, it must be
adaptable to any environment and agent architecture; in other words,
the system must be independent of the environment. Some principles of
general intelligence are depicted in
\prettyref{fig:general-principles}.

\section{Architecture Overview}

Many of the aforementioned problems have been addressed by existing AI
architectures.  We therefore take a well-understood general AI
architecture called the Alpha architecture of Solomonoff
\cite{solomonoff-progress}, and define some basic capabilities better,
while incorporating newer models and methods from recent research.

For the purposes of general-purpose AI, two most significant events
have occured since Alpha was designed in 2002. First, the Gödel
Machine architecture \cite{Schmidhuber:05gmai} which also provides a
level of self-reflective thinking, and presents an agent model around
it. The other notable development is the immense success of deep
learning methods, which now enable machines to achieve pattern
recognition at human-level or better for many basic tasks.  The
present design therefore merges these two threads of developments into
the Alpha framework. The architecture also provides for basic
universal intelligent agents, and self-reflection like Gödel Machine
does. Unlike Gödel Machine, we do not assume that the environment is
known to a substantial degree, such things are assumed to be learnt.

Like the Alpha architecture, we assume a basic problem solver that is
smart enough to bootstrap the rest of the system. This component is
called the AI Kernel.

The system is thought to be parameter free, dependent only on the
data, and the commands given. The system's interface is a graphical
web-based application that allows the user to upload datasets and then
apply AI tasks from the library. The system also provides an API for
programming novel tasks. A basic graphical programming environment is
considered for later releases since the system aims to be usable by
non-programmers.

\subsection{Components}

We review the major components of the system architecture, and explain
their functions.

\subsubsection{AI Kernel}

The AI kernel is an inductive programming system that should use a
universal reference machine such as LISP. We have proposed using
Church as the reference machine of such a system.  However, what
matters is that the AI kernel must be able to deal with all types of
data, and tasks. We assume that the reference machine is variable in
the right AI kernel. The kernel must be a compact code base that can
run on a variety of hardware architectures to ensure portability, and
the parallelization must support heterogeneous supercomputing
platforms for high energy efficiency and scalability.

The AI kernel supports sophisticated programmability, allowing the
user to specify most machine learning tasks with a very short API.  We
employ OCaml generic programming to characterize the kernel's internal
components, model discovery, and transfer learning algorithms.

The AI kernel supports real-time operation, and can be configured to
continuosly update long-term memory splitting running-time between
currently running task and meta-learning.

\subsubsection{Bio-mimetic Search}

State-of-the-art bio-mimetic machine learning algorithms based on such
methods as stochastic gradient descent, and evolutionary computation
are available in the AI kernel, and thus chosen and used
automatically.

\subsubsection{Heuristic Algorithmic Memory 2.0}

The AI kernel has integrated multi-term memory, meaning that it solves
transfer learning problems automatically, and can remember solutions
and representational states at multiple time scales.  Heuristic
Algorithmic Memory 2.0 extends Heuristic Algorithmic Memory
\cite{DBLP:conf/agi/Ozkural11} to support multiple reference machines.

\subsubsection{Problem Solvers}

Problem Solution Methods (PSMs) are methods that solve a given
problem.  These could be algorithmic solutions like sorting a list of
numbers, or statistical methods like predicting a variable. The Alpha
architecture basically tries a number of PSMs on a problem until it
yields. However, in Omega, it is much better specified which PSMs the
system should start with.  Since the system is supposed to deal with
unknown environments, we give priority to machine learning and
statistical methods, as well model classes that directly address some
challenging properties of the environment, and support hard
applications like robotics. The diversity of the model classes and
methods supported expand the range of Omega applications. The Alpha
architecture can invent and retain new PSMs, that is why it should be
considered an open-ended architecture; so is Omega.

The architecture is taught how to use a problem solver via
unstructured natural language examples, like the intent detection task
in natural language processing.

Both narrowly specialized and general-purpose methods are included in
the initial library of problem solvers for initially high
machine-learning capability.

For approximating functions, there are model-based learning algorithms
like a generic implementation of stochastic gradient for an arbitrary
reference machine. For model discovery, model-free learning algorithms
like genetic programming are provided. Function approximation
facilities can be invoked by the ensemble machine to solve machine
learning problems.  Therefore, a degree of method independence is
provided by allowing multi-strategy solvers.

A basic set of methods for solving scientific and engineering problems
is provided. For computer science, the solutions of basic algorithmic
problems including full software development libraries for writing
basic computer programs for each reference machine (standard library).
For engineering, basic optimization methods and symbolic algebra. In
the ultimate form of the architecture, we should have methods for
computational sciences, physical, and life sciences.

A full range of basic data science / machine learning methods are
provided including:
\begin{description}
\item[Clustering] Clustering is generalized to yield automated
  statistical modeling. Universal induction can be used to infer a PDF
  minimizing expected divergence (AI kernel function). Both
  general-purpose and classical clustering algorithms are provided, in
  recognition that for a specific class of problems a specialized
  method can be faster, if not necessarily more accurate. The
  classical algorithms of k-means \cite{hartigan1979algorithm},
  hierarchical agglomerative clustering \cite{murtagh2014ward}, and
  Expectation Maximization (EM) for Gaussian Mixture Models
  \cite{xu1996convergence} are provided. General-purpose algorithms
  based on NID \cite{vitanyi2009normalized}, and universal induction
  enable working with arbitrary domains.
\item[Classification] Again both classical and general-purpose
  algorithms are supported. Classical algorithms of decision-tree
  classifier \cite{quinlan2014c4}, random forest
  \cite{liaw2002classification}, knn \cite{altman1992introduction},
  logistic regression \cite{cox1958regression}, and SVM
  \cite{vapnik1995support} are supported. General-purpose algorithm
  invokes AI kernel universal induction routines to learn a mapping
  from the input to a finite set. NID based classifier works with
  arbitrary bitstrings.
\item[Regression] General-purpose algorithm invokes universal
  induction routines in the AI kernel to learn a stochastic operator
  mapping from the data domain to a real number. Classical algorithms
  of linear regression, logistic regression, and SVM are supported.
\item[Outlier detection] The generalized outlier detection finds the
  points least probable given the rest of the dataset using a
  generalization of z-score; to first model the data again a universal
  set induction invocation characterizes the data.
\item[Time-series Forecasting] Time series prediction is generalized
  with a universal induction approach modeling the stochastic
  dynamics, then the most probable model is inferred. Classical
  time-series prediction algorithms of ARIMA, Hidden Markov Model
  (HMM), and Hiearchical Hidden Markov Model (HHMM)
  \cite{fine1998hierarchical} are provided. A deep LSTM based forecast
  method is also provided \cite{Hochreiter:97nips}.
\item[Deep Learning] A complete range of DNN architectures for various
  data types such as image, audio, video and text are provided.
  Standard algorithms of backpropagation, stochastic gradient and
  variational inference are supported. The state-of-the-art fully
  automated machine learning algorithm of Fourier Network Search (FNS)
  \cite{koutnik2014gecco} is included.  We also invoke universal
  induction routines to automate neural model discovery. The deep
  learning implementations are parallelized for multi GPU
  clusters. For this purpose, an existing deep learning framework such
  as TensorFlow may be used. The deep learning framework we use is a
  different, proprietary approach that predates TensorFlow and is
  composed of a neural programming language called MetaNet and a
  heterogeneous supercomputing middleware called Stardust.
\end{description}

Each algorithm mentioned is exposed as a PSM in the system.

\subsubsection{Ensemble Machine}

An ensemble machine is introduced to the system which runs PSMs in
parallel with time allocated in accordance with their expected
probability of success. The associations between tasks and their
success are remembered as a stochastic mapping problem solved with the
universal induction routines of AI Kernel, guiding future
decisions. The ensemble machine itself is exposed as a PSM.

\subsection{Representation Languages}

We define eight reference machines to widen the range of solutions
obtainable, and types of environments/applications addressable.

\begin{description}
\item[MetaNet] MetaNet is a new General Neural Networks (GNN)
  representation language that encompasses common neuron types and
  architectures used in neural network research.  It is a graphical
  meta-language that can be used to define a large number of network
  architectures. Formally, it uses a multi-partite labeled directed
  graph with typed vertices, as a generic representation to represent
  neural circuits, and the richer sort of representation allows us to
  extend the model to more biologically plausible, or with
  neuroscience-inspired models.  The system uses this representation
  to facilitate automated model discovery of the right neural network
  for the given task when evaluating the MetaNet representation
  language.

\item[Church] We use the Church language \cite{goodman2012church} to
  represent probability distributions and solve basic algorithmic
  problems like adding a list of numbers, and the Towers of Hanoi
  problem. Components expose their interfaces in Church machine,
  expanding self-reflection capability.

\item[Probabilistic Logic] We define a probabilistic logic programming
  language to deal with uncertainty and stochasticity, and the ability
  to solve reasoning problems.

\item[Bayesian Networks] We define a general class of bayesian
  networks that can be used to deal with uncertainty.

\item[Analog Computer] We use an analog computing model to represent
  dynamical, continuous and stochastic systems better.

\item[Picture] We use the Picture \cite{kulkarni2015picture} language
  to deal with images.

\item[Matrix Computer] We use an LAPACK based matrix algebra
  computing package such as GNU Octave to represent mathematical
  solutions.

\item[Asynchronous Computer] We define an asynchronous model of
  computation for conception of fine-grain concurrent models.

\end{description}

\subsection{Neural Representation Classes}

There are a number of ready neural representations that the system can
quickly invoke.

\begin{description}
\item[Fourier Neural Network] Fourier Neural Networks use a Fourier
  series representation to represent neural networks compactly, and
  may be considered a general-purpose learning model class
  \cite{koutnik:agi10,koutnik:gecco10}.
\item[Convolutional Neural Networks] CNNs are particularly effective
  for pattern recognition problems. A variety of basic CNNs
  \cite{lecun1995convolutional} suitable for processing different
  kinds of data are provided, including specialized networks such as
  multi-column DNNs for image classification \cite{ciresan:2012NN},
  for video \cite{karpathy2014}, text \cite{zhang15}, and speech
  \cite{abdel2014convolutional}.
\item[Deep Belief Networks] These networks are a stack of Restricted
  Boltzmann Machines \cite{hinton:06afast} that can perform
  unsupervised learning.
\item[Deep Autoencoders] Deep autoencoders \cite{hinton2006reducing}
  use several hidden unit layers, two deep belief networks, that learn
  to compress and then reproduce the data. We provide specific
  applications like variational autoencoders for image captions
  \cite{pu2016}, inverse graphics \cite{kulkarni15}, multimodal
  learning \cite{ngiam2011multimodal}.
\item[LSTM/GRU networks] We provide a variety of RNN models using LSTM
  (Long-Short Term Memory) and GRU (Gated Recurrent Unit) stacks to
  model sequential data. Variants for different data types such as
  speech \cite{graves2014}, video \cite{Zhang2016VideoSW}, image
  \cite{lrcn2014} are included.
\item[Recursive Deep Networks] Especially useful for language
  processing, these networks can recognize hierarchical structures
  easily \cite{socher2013recursive}.
\end{description}

The networks are specified as generic network architectures that can
scale to required input/output size.  Any hyper-parameters are
designated as variables to be learned to the AI kernel so that the
hyper parameters can adapt to the problem.  These networks are
considered to be sufficient as providing enough library
primitives. The generators for neural networks are specified such that
the program generator can indeed generate all of the library networks;
however, re-inventing the wheel is not a feasible idea, therefore we
aim to include a complete inventory of deep learning models.

\subsection{Software Architecture}

\subsubsection{Functional Decomposition}

\begin{figure*}
  \centering
  \includegraphics[width=\textwidth]{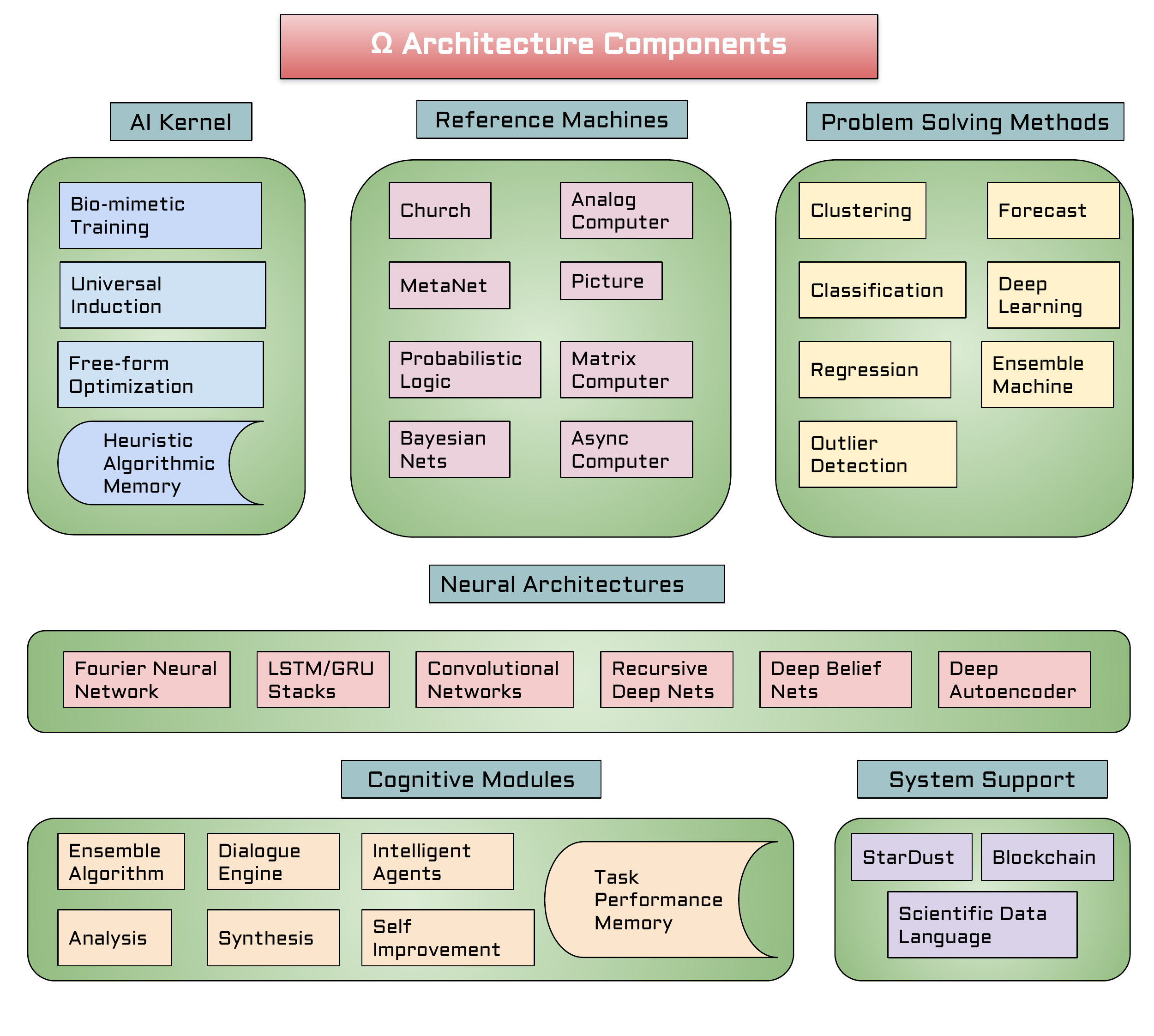}
  \caption{$\Omega$ Architecture Component Diagram}
  \label{fig:omegaarch}
\end{figure*}

A high-level component architecture without the many inter-component
interactions is depicted in \prettyref{fig:omegaarch}.

The system's process flow is straightforward. The user presents the
system with a number of datasets, and the user selects a task to be
applied to the data. The system automatically recognizes different
data types, however, it also allows data to be specified in detail by
a description language. The system will also accept tasks to be
defined via a conversational engine, and a programming interface
(API). The conversational engine can learn to recognize a task via
given examples, mapping text to a task specification language and
backwards. The programming interface accumulates the interfaces of all
the components, unified under a single facade of a generic problem
solver, which is formulated as a general optimizer \cite{alpcan14}. As
in Alpha, the most general interface the system provides is that of
time-limited optimization, however, the system allows to solve any
well-defined problems allowing the user to define any success
criterion.  The problem solver then predicts the probability that a
PSM will succeed in solving the so-specified problem, and then
translates the input data and the task to a format that the particular
PSM will understand, and also translate any results back. After a task
is solved, the system automatically updates its long-term memory and
writes a snapshot to the disk. It then executes higher-order cognition
routines to improve its PSMs, and awaits for the next task.

\subsubsection{Execution}

The execution of PSMs is parallelized as much as possible, as many
PSMs may be run in parallel, but also some methods will allow data to
be sharded, and will also parallelize well themselves. A main
operational goal of the system is the ability to keep track of these
parallelizations well enough to present an OS like stability to the
user with a simple interface.  The system also allows modules to be
invoked concurrently and in a distributed manner to facilitate the
design of distributed and decentralized applications using the API.

The PSMs are executed with a hardware abstraction layer called
Stardust that provides heterogeneous peer-to-peer computing capability
to the architecture.  MetaNet acts as a common neural network
representation language. Scientific Data Language is a data
specification language that allows us to describe the type, format and
semantic labels of the data.

\subsection{Higher-Order Cognition}

Two fundamental higher-order cognitive functions are defined as
analysis and synthesis. Analysis decomposes a problem into components
and then tries to solve the problem by first solving sub-problems and
then merging their results into a solution. Synthesis generates new
PSMs by combining known PSMs. These operations give the ability to
observe the code of its modules, and expand the system's repertoire of
PSMs continuously.  Analysis is self-reflective in that sense, and
synthesis is self-reification.

These functions correspond to a second kind of modularity where the
tasks themselves can be decomposed, and entirely new PSMs may be
invented and added as new modules to the system.

The system continually self-reflects through updating its algorithmic
memory for accelerating future solutions. It also keeps a record of
task performance for trying to retroactively optimize past solutions.
The components expose themselves via a high-level reference machine
(Church) which acts as the system ``glue code'' to compose and
decompose system functions. Since Church is quite expressive, it can
also act as the system's task description code, and be used to
recognize, decompose, and compose tasks and solutions. The synthesis
and analysis modules operate over the system's modular cognition
itself, helping with synthesis of new solution methods and analysis of
problems. The system uses self-models to guide its self-improvement,
for instance, by trying to optimize its performance.

\subsection{Self-Improvement}

Analysis and synthesis can learn how to accomplish this as they can
use the execution history to improve the results
retrospectively. After a new problem is solved, therefore, the system
can continuously try to improve its consolidated memory of PSMs by
trying to generate new PSMs that will improve performance over
history, or by decomposing problems to accelerate their execution. A
general objective such as maximizing energy efficiency of solutions
can be sought for self-improvement.

\subsection{Modular Neural Architectures}

PSMs embody a basic kind of modularity in the system which are
extended with modular neural architectures. These architectural
schemas are a cortical organization that decomposes the networks into
many cortical columns, which are henceforth again decomposed into
micro-columns, with variant geometries. This organization schema is
called MetaCortex, and it is a way to describe larger networks that
can digest a variety of data sources, and construct larger neural
models with better modularity, that is better data/model encapsulation
based on affinity. There are architectures such as multi-column
committee networks that already implement these architectures,
however, we would expand this to the entire library of networks
described.

\subsection{Intelligent Agents}

Basic goal-following and utility-maximization agents can be realized
similarly to time-series prediction. A typical two part model of
learning representations (world model), and planning will be
provided. A basic neural template will provide for multi-modal
perception, multi-tasking, task decomposition and imitation learning.
Neural templates corresponding to different kinds of agents such as
Deep Mind's I2A model \cite{weber2017imagination} will be provided.

The intelligent agents have a real-time architecture, they run at a
fixed number of iterations every second. At this shortest period of
synchronization, mostly backpropagation like learning algorithms, and
simulation are allowed to complete.  Everything else is run in the
background for longer time-scales.

\subsection{Process and Memory Hierarchy}

The processes and memory are organized hierarchically from long-term,
heavy tasks to short-term, lightweight tasks. At the shortest scale,
the system has neural memory units like LSTM, that last at the scale
of one task, and model-based local training/inference algorithms like
backpropagation algorithms. At a longer scale which corresponds to one
iteration of problem solution procedure, the system remembers the best
solutions so far, and it updates its mid-term memory with them to
improve the solution performance in the next iteration. At this scale,
the system will also engage in more processes such as the just
mentioned memory update operation, and more expensive training
algorithms such as genetic algorithms.  At the highest scale, the
system runs the most expensive model-free learning algorithms that can
search over architectures, models, and components, and updates its
persistent, long term memory based on the statistics about solutions
of the new problem after solving it to guide the solution of new
problems. The system also updates its PSMs by executing its
higher-order cognitive functions at this scale.

\subsection{Hardware Abstraction Layer}

The architecture depends on a Hardware Abstraction Layer (HAL) in the
form of Stardust peer-to-peer computing substrate. Stardust provides a
bytecode representation that can be run on both multi-core cluster,
GPU clusters, and FPGA clusters in the future. Stardust uses
virtualization technology for compartmentalization and basic security.
It uses a lightweight kernel, and provides parallel and distributed
computing primitives.

\subsection{Peer-to-Peer Computing}

Peer-to-peer computing is facilitated by a node software that users
download and operate to earn fees from the network with a
cryptographic utility token. Since approaching human-level will
typically require several petaflops/sec of computing speed, scaling to
a significant number of global users requires peer-to-peer
computing. If a proportion of profit is paid to the users, this can
incentivize their contribution, providing a cost-effective computing
platform for the architecture.

\subsection{Scientific Data Language}

A canonical data representation seems essential for Alpha family of
architectures, because the PSMs can vary wildly in their assumptions.
That is why, a common data format is required. The format we propose
has standard representations for both structured (tabular, tree,
network, etc.), unstructured (like text, audio, image, video) and
complex data types. It supports web, cloud and fog computing data
sources, thus also abstracting data ingestion.  Each PSM handles the
data differently, mapping to an internal representation if
necessary. Therefore, every element of the data can be given a type,
the format of the data may be specified (such as a 10x10 table of
integers), and semantic labels may be ascribed to data elements. For
instance, each dataset has a different domain, which may be designated
with a domain path.  Likewise, the physical units, or other semantic
information may be annotated on a dataset with arbitrary
attributes. The specification language must be also modular allowing
to include other modules.  The system automatically ingests known data
formats, recognizes them, and converts into this common format which
may henceforth be modified by the user. In the future, we are planning
to design a data cleaning facility to improve the data at this stage.

% \section{Cloud Robotics Use Case}

\section{Discussion and Research Program}

We gave the overview of an ambitious architecture based on
Solomonoff's Alpha Architecture, and Schmidhuber's Gödel Machine
architecture. The system is like Alpha, because it re-uses the basic
design of PSMs. It is also similar to Gödel Machine architecture,
because it can deploy a kind of probabilistic logical inference for
reasoning and it can also observe some of its internal states and
improve itself. The system also has basic provisions for intelligent
agents, but it is not limited to them. We saw that the first important
issue with implementing Alpha was to decide a basic set of primitives
that will grant it sufficient intelligence to deal with human-scale
problems. It remains to be demonstrated empirically that is the case,
however, two of the eight reference machines have been implemented and
seen to operate effectively.

A criticism may be raised that we have not explained much about how
the AI Kernel works. We only assume that it presents a generalized
universal induction approximation that can optimize functions, rich
enough to let us define basic machine learning tasks. It surely cannot
be Levin search, but it could be any effective multi-strategy
optimization method such as evolutionary architecture search
\cite{liang2018evolutionary}.  We are using an extension of the
approach in Fourier Network Search \cite{koutnik:gecco10} which is
also likely general enough. The memory update is also not detailed but
it is assumed that it is possible to extend an older memory design
called heuristic algorithmic memory so that it works for any reference
machine. We also did not explain in detail how many components work
due to lack of space, which is an issue to be tackled in a longer
future version of the present paper.

In the future, we would like to support the architectural design with
experiments, showing if the system is imaginative enough to come up
with neural architectures or hybrid solutions that did not appear to
humans. The algorithms used are expensive, therefore they might not
work very well with the extremely large models required by the best
vision processing systems; but to accommodate such models, it might be
required that the system evolves only parts of the system and not the
entire architecture. The system is intended to be tested on basic
psychometric tests first, and a variety of data science problems to
see if we can match the competence of the solution a human data
scientist would achieve.

\bibliographystyle{named}
\bibliography{agi,deep,ann,neuro,datascience}

\end{document}